# Novel Metric based on Walsh Coefficients for measuring problem difficulty in Estimation of Distribution Algorithms

Saeed Ghadiri
*K.N. Toosi University of technology*
saeed.ghadiri@alumni.kntu.ac.ir

Amin Nikanjam
*K.N. Toosi University of technology*
nikanjam@kntu.ac.ir

*Abstract—* Estimation of distribution algorithms are evolutionary algorithms that use extracted information from the population instead of traditional genetic operators to generate new solutions. This information is represented as a probabilistic model and the effectiveness of these algorithms is dependent on the quality of these models. However, some studies have shown that even multivariate EDAs fail to build a proper model in some problems. Usually, in these problems, there is intrinsic pairwise independence between variables. In the literature, there are few studies that investigate the difficulty and the nature of problems that can not be solved by EDAs easily. This paper proposes a new metric for measuring problem difficulty by examining the properties of model-building mechanisms in EDAs. For this purpose, we use the estimated Walsh coefficients of dependent and independent variables. The proposed metric is used to evaluate the difficulty of some well-known benchmark problems in EDAs. Different metrics like Fitness Distance Correlation (FDC) are used to compare how well the proposed metric measures problem difficulty for EDAs. Results indicate that the proposed metric can accurately predict the EDA's performance in different problems.

## I. Introduction

Problem hardness is studied in the field of evolutionary computation for over two decades. Usually, a problem is classified as "easy" if it can be solved by an algorithm in computationally linear runtime [1]; and is classified as "hard" the algorithm needs exponential runtime. It has long been a challenge for researchers to classify fitness functions as "easy" or "hard". A fitness function can be difficult for one algorithm, but easy for another. For instance, some EAs might find a unimodal function difficult, while others may find it easy. An EA can have difficulty with a non-deceptive function, while it may have no problems with a deceptive one. So a problem's hardness is influenced by the algorithm employed. [2].

There are many optimization problems that evolutionary algorithms can solve effectively, but choosing the best algorithm for a given problem is a challenge. Traditional methods of measuring the difficulty of problems are independent of algorithms; they utilize features from the problem space to determine the difficulty of problems for all evolutionary algorithms.

This paper aims to Introduce a metric that measures problem difficulty while taking into consideration the features of the EDAs. A major concern in EDAs is linkage learning, which involves estimating joint probabilities and constructing models by using linkage information. A problem becomes challenging for EDAs when extracted statistics from population shows incorrect correlation between problem variables; The dependent variables can either show no correlation or similar statistics to the independent variables. We thus propose a metric for EDA problem difficulty that compares the statistics between dependent variables and independent variables of the problem.

The rest of this paper is organized as follows: Section II briefly reviews the necessary background knowledge about EDAs and describes some concepts of schema theory and problem difficulty which are used along the paper. Section III discusses linkage learning problem for EDAs from an information theory perspective. Section IV reviews Walsh analysis and introduces Walsh coefficent as a measurement for problem difficulty in linkage learning. The results of the experiments using proposed metric are discussed in section V and their meaning is explained. In Section VI, the paper is summarized and some conclusions are drawn.

## II. Background

### A. Estimation of Distribution Algorithms

Estimation of distribution algorithms (EDAs) are evolutionary optimization algorithms that build probabilistic models with information extracted from a population of promising solutions. The built model encodes the dependencies between the variables of the problem and is later used to generate new solutions by model-guided crossover or sampling. Since Estimation of Distribution Algorithms (EDA) identify dependencies between variables, they were employed successfully to solve problems that are considered hard for other EAs.

EDAs differ in terms of the model they deploy to capture dependencies, as well as the methods they employ to learn and sample the model. The earliest EDAs assumed independence between the variables of the problem and models were just a probability vector, like population-based incremental learning (PBIL) [3], compact genetic algorithm (cGA) [4], and univariate marginal distribution algorithm (UDMA) [5]. There are often dependencies between variables in many problems, making it impossible for these simple models to represent such interactions. More complex EDAs with bivariate factorization like mutual information maximizing input clustering (MIMIC) [6], bivariate marginal distribution algorithm (BMDA) [7] can capture these interactions. But some problems have multivariate dependencies so multivariate factorization models like extended compact genetic algorithm (eCGA) [8] and Bayesian optimization algorithm (BOA)[9] were proposed.

*B. Schema Theory and Linkage learning*

Evolutionary algorithms have a step called selection in which the best individuals from the population are chosen, and these individuals are used for reproduction. During the selection phase, EA selects a similar feature among the best individuals which is called schema.[10] Reproduction operators such as mutation and crossover should not disrupt the schema to help the algorithm progress over generations[11]. A building block (BB) is a schema consisting of solutions with higher fitness values than the population's average fitness. The building block hypothesis suggests that EA performs best when it uses these blocks effectively.

EDAs find bivariate and multivariate dependencies from samples drawn population and build the structure of a probabilistic graphical model which is similar to a linkage model. A linkage model is a matrix showing dependencies between problem variables. Based on the ideas of Holland [10], the concept of Building Block is connected with statistical dependencies. So reproduction operators may produce unpromising solutions if they disrupt previously found dependencies and their success depends on the accuracy of the models they use. So linkages found by the model play a key role in the algorithm's success.

*C. Fitness landscape and problem difficulty*

Fitness function is not sufficient to represent the structure of a problem as it does not relate solutions to one another. So in order to understand how and where heuristic algorithms operate, fitness landscapes were introduced[12].

The set of local optima in a fitness landscape can be used to define how hard a problem is. If a distribution has only one global optimum, then it is called unimodal, while if it has multiple global optimums, then it is called multimodal. Distribution of local optimums in a problem can be informative about the fitness landscape. In [13] the distances between local optimums were studied to answer the question of whether better optimums are near each other or not. If fitness value increases in an area crowded with local optimums, a problem is easy.

Another metric to measure problem hardness is ruggedness[14] which is the correlation of fitness values along a random walk in the landscape. This means simply counting the number of local optima in the slope from uphill to downhill. Neutrality measures problem hardness as connected regions with similar fitness in a landscape[15]. Fitness distance correlation(FDC) is the most commonly used metric as a measure of difficulty[16]. In this method samples from the fitness landscape are drawn and the correlation between fitness values and distance to global optimum is calculated. A problem is called deceptive or hard if this metric shows a high negative correlation. Another analysis for fitness landscape is called spectral landscape analysis. One method uses the Fourier series to transform the landscape into a set of simpler landscapes and make the analysis easier[17].

*D. Additively separable fitness functions*

A pseudo-boolean fitness function inputs a chromosome of $n$ binaries and outputs a real number

$$f: \{0,1\}^n \rightarrow \mathbb{R} . \qquad (1)$$

These fitness functions are designed to investigate and compare evolutionary algorithms. Additively separable functions consist of a set of different subfunctions in which the optimal solution to the problem can be identified

$$f(x) = \sum_{i=1}^{m} f_i(x, \pi_i), \ x \in \{0,1\}^n , \qquad (2)$$

where $\pi_i$ is the hyperplane for subfunction $i$. Each subfunction is separate from the others $\bigcap_{i=1}^{m} \pi_i = \emptyset$. These fitness functions are representative of real-world problems that can be solved by divide and conquer methods. An accurate linkage learning method should be able to discover the structure between subfunctions as well as how they contribute to the total fitness value. A group of additively separable functions called deceptive fitness functions are designed to challenge Evolutionary algorithms.

### III. PROBLEM DIFFICULTY FOR EDAS

The most important step in bivariate and multivariate EDAs is linkage learning, where the PGM is built on observed statistical dependencies in the population. So a difficulty metric could be defined in the linkage learning step. In most EDAs, the main issue is incorrect correlation or fake dependency in the problem. This problem arises when populations show dependencies not presented in the underlying problem. Fake dependency can be solved by increasing the population size, but in some problems, this method does not work.

Learning how to build accurate models is crucial for EDAs to solve problems successfully. However, empirical results have shown that learning more complex models will not always increase the efficiency of EDAs [18]. For example, BOA requires an exponential number of fitness evaluations to solve separable Concatenated Parity functions, while CGA and eCGA can find the optimum in polynomial time [19]. According to the authors, The model constructed by BOA consists of spurious linkages that mislead the search into non-promising areas. So a complex model is not necessarily more accurate than a simple model. on the other hand increase in the degree of interactions between variables can make extracting accurate information from the population problems really hard and as a result, can make the problem difficult for EDAs. So in those cases, simpler models like single variable and bivariate EDAs perform better than more complex models in some NK-Model problems.

assume that $f: \{0,1\}^n \rightarrow \mathbb{R}$ is an additively separable function which can be divided into $m = (\frac{N}{K})$ subfunctions $f_\pi$ and $\mathscr{F}$ is set of disjoint subset of variables assigned to subfunctions. If $\mathscr{F} = \{\pi_1, \pi_2\}$ that separates the fitness function into $f(x) = f_{\pi_1}(x_{\pi_1}) + f_{\pi_2}(x_{\pi_2})$, then $X$ and $Y$ are random variables in $\pi_1$ and $Z$ is a random variable in $\pi_2$ [20]. Thus $X$ and $Y$ are dependent and $X$ and $Z$ are independent of each other.

So EDA can successfully recognize statistical dependency between X and Y if

$$I(X;Y) > I(X;Z),\qquad(3)$$

Where $I(X;Y)$ is the mutual information between $X$ and $Y$. So we can write

$$H(X) + H(Y) - H(X,Y) > H(X) + H(Z) - H(X,Z)$$
$$H(Y) - H(X,Y) > H(Z) - H(X,Z),\qquad(4)$$

Where $H(X)$ is the entropy of $X$ and $H(X,Y)$ mutual entropy between $X$ and $Y$. In the simplest case where all subfunctions are similar, $H(Y)$ and $H(Z)$ are equal

$$H(X,Y) < H(X,Z)).\qquad(5)$$

In linkage learning methods using bivariate statistics, these conditions are essential for learning accurate models. In a multimodal problem, the distribution of $(X,Y)$ can be uniform so the entropy is high and as a result, no mutual information can be found, leading the learning algorithm to believe that X and Y are independent. So based on equation (3) and (5) two metrics can be defined as[21]

$$\delta_I = I(X;Y) - I(X;Z)\qquad(6)$$

$$\delta_H = 1 - H(X,Y)/H(X,Z).\qquad(7)$$

## IV. WALSH COEFFICIENTS FOR PROBLEM DIFFICULTY

### A. Walsh Analysis

Another method that can simplify the fitness landscape of a pseudo-boolean function is Walsh decomposition which transforms the function into Walsh coefficients. For a binary problem in $\mathbb{R}^{2^n}$ space, where $n$ is the length of the problem, the set of Walsh coefficients is a vector with size of $2^n$

$$\alpha_n = \begin{bmatrix} \alpha_\emptyset \\ \alpha_{\{0\}} \\ \alpha_{\{1\}} \\ \alpha_{\{0,1\}} \\ \alpha_{\{2\}} \\ \alpha_{\{0,2\}} \\ \alpha_{\{1,2\}} \\ \alpha_{\{0,1,2\}} \\ \vdots \\ \alpha_{\{0,1,2,\ldots,n-1\}} \end{bmatrix}.\qquad(8)$$

For example using Walsh-Hadamard transform[22], for the function $f(x) = 3 + x_0 - 2x_1$ the Walsh coefficients are

$$\alpha_n = \begin{bmatrix} \alpha_\emptyset \\ \alpha_{\{0\}} \\ \alpha_{\{1\}} \\ \alpha_{\{0,1\}} \end{bmatrix} = \begin{bmatrix} 2.5 \\ 0.5 \\ -1.0 \\ 0.0 \end{bmatrix}.\qquad(9)$$

So the function can be written as $2.5\,W_\emptyset(x) + 0.5\,W_{\{0\}}(x) - W_{\{1\}}(x)$ which indicates that $X_0$ and $X_1$ are independent in this function. Walsh coefficients can be calculated using $2^n \times 2^n$ Hadamard matrix defined as

$$H.H^T = 2^n I\qquad(10)$$

$$H_n = \begin{bmatrix} H_{n-1} & H_{n-1} \\ H_{n-1} & -H_{n-1} \end{bmatrix}\qquad(11)$$

$$H_1 = \begin{bmatrix} 1 & 1 \\ 1 & -1 \end{bmatrix},\qquad(12)$$

So Walsh decomposition is defined as

$$\alpha_n = \frac{1}{2^n} H_n \mathbf{f}_n.\qquad(13)$$

For the function $f(x) = 3 + x_0 - 2x_1$, Walsh decomposition is

$$\mathbf{f}_n = \begin{bmatrix} f_{11} \\ f_{01} \\ f_{10} \\ f_{00} \end{bmatrix} = \begin{bmatrix} 2 \\ 1 \\ 4 \\ 3 \end{bmatrix}\qquad(14)$$

$$\alpha_n = \frac{1}{2^n} H_n \mathbf{f}_n = \frac{1}{4}\begin{bmatrix} 1 & 1 & 1 & 1 \\ 1 & -1 & 1 & -1 \\ 1 & 1 & -1 & -1 \\ 1 & -1 & -1 & 1 \end{bmatrix}\begin{bmatrix} 2 \\ 1 \\ 4 \\ 3 \end{bmatrix} = \begin{bmatrix} 2.5 \\ 0.5 \\ -1.0 \\ 0.0 \end{bmatrix}.$$

Fast Walsh-Hadamard[23] is a method with fewer calculations for the transform which can decompose the function in $O(n\log n)$ instead of $O(n^2)$.

So by using Walsh decomposition, we can simplify the fitness landscape to perform our analysis on it. The average fitness value of a schema can be calculated by the corresponding Walsh coefficients as

$$f(H) = \alpha_\emptyset + \sum_{j \subset J} \alpha_j \cdot (-1)^{u(j)},\qquad(15)$$

where $H$ is the schema, $J$ is the subset of schema that is determined, $\alpha$ is the Walsh coefficient and $u(j)$ is the number of ones in $J$. In table 1 this calculation is performed for a problem with $n = 3$ by different schemas. It is important to note that schemas with higher fitness values are built with more elements whereas lower fitness schemas are built with fewer elements.

TABLE I. FITNESS VALUE FOR SCHEMAS IN A PROBLEM WITH N=3

| Average Fitness Value by Walsh Coefficient | Schema |
|---|---|
| $\alpha_\emptyset$ | *** |
| $\alpha_\emptyset + \alpha_{\{2\}}$ | **0 |
| $\alpha_\emptyset - \alpha_{\{2\}}$ | **1 |
| $\alpha_\emptyset - \alpha_{\{0\}}$ | 1** |
| $\alpha_\emptyset + \alpha_{\{1\}} + \alpha_{\{2\}} + \alpha_{\{1,2\}}$ | *00 |
| $\alpha_\emptyset + \alpha_{\{1\}} - \alpha_{\{2\}} - \alpha_{\{1,2\}}$ | *01 |
| $\alpha_\emptyset - \alpha_{\{1\}} - \alpha_{\{2\}} + \alpha_{\{1,2\}}$ | *11 |
| $\alpha_\emptyset - \alpha_{\{0\}} - \alpha_{\{1\}} + \alpha_{\{0,1\}}$ | 11* |
| $\alpha_\emptyset + \alpha_{\{0\}} + \alpha_{\{1\}} - \alpha_{\{2\}} + \alpha_{\{0,1\}} - \alpha_{\{0,2\}} - \alpha_{\{1,2\}} - \alpha_{\{0,1,2\}}$ | 001 |
| $\alpha_\emptyset - \alpha_{\{0\}} + \alpha_{\{1\}} - \alpha_{\{2\}} - \alpha_{\{0,1\}} + \alpha_{\{0,2\}} - \alpha_{\{1,2\}} + \alpha_{\{0,1,2\}}$ | 101 |

A non-zero coefficient on a clique of more than one variable shows the interaction between the variables in that clique[22]. For example, If there were no bivariate interaction between $A$ and $B$, the Walsh coefficient $\alpha_{\{A,B\}}$ would be zero. Analysis of interactions between variables of the problem can be performed using these coefficients. Walsh coefficients can also be estimated from average fitness values of different schemas. For a problem with $n = 3$, based on equation (13) the coefficients are calculated as

$$\frac{1}{8}\begin{bmatrix} 1 & 1 & 1 & 1 & 1 & 1 & 1 & 1 \\ 1 & -1 & 1 & -1 & 1 & -1 & 1 & -1 \\ 1 & 1 & -1 & -1 & 1 & 1 & -1 & -1 \\ 1 & -1 & -1 & 1 & 1 & -1 & -1 & 1 \\ 1 & 1 & 1 & 1 & -1 & -1 & -1 & -1 \\ 1 & -1 & 1 & -1 & -1 & 1 & -1 & 1 \\ 1 & 1 & -1 & -1 & -1 & -1 & 1 & 1 \\ 1 & -1 & -1 & 1 & -1 & 1 & 1 & -1 \end{bmatrix}\begin{bmatrix} f_{111} \\ f_{011} \\ f_{101} \\ f_{001} \\ f_{110} \\ f_{010} \\ f_{100} \\ f_{000} \end{bmatrix} = \begin{bmatrix} \alpha_{\{\emptyset\}} \\ \alpha_{\{0\}} \\ \alpha_{\{1\}} \\ \alpha_{\{0,1\}} \\ \alpha_{\{2\}} \\ \alpha_{\{0,2\}} \\ \alpha_{\{1,2\}} \\ \alpha_{\{0,1,2\}} \end{bmatrix}. \quad (16)$$

So the coefficient $\alpha_{\{1,2\}}$ is

$$\begin{aligned}\alpha_{\{1,2\}} &= \frac{1}{8}\big((f_{111}+f_{011})-(f_{101}+f_{001})\\ &\quad -(f_{110}+f_{010})+(f_{100}+f_{000})\big) \quad (17)\\ &= \frac{1}{4}((f_{*11}-f_{*01}-f_{*10}+f_{*00}).\end{aligned}$$

Thus using estimated fitness values of $f_{*11}$, $f_{*01}$, $f_{*10}$ and $f_{*00}$ the coefficient between the second and third bit can be calculated. This equation is similar to the non-linearity metric used by the LINC algorithm[24] to find linkages in the model

$$m_{ij} = |f_{a_i=0,a_j=1}+f_{a_i=1,a_j=0}-f_{a_i=0,a_j=0}-f_{a_i=1,a_j=1}|. \quad (18)$$

So in another way the Walsh coefficients can be estimated from fitness value of schemas as

$$\alpha_M = \frac{1}{2^{u(m)}}\sum_{i \in B_m}(-1)^{u(i)}.f(i \oplus *), \quad (19)$$

Where $M$ is a subset of variables, $m$ is mask of the subset, $u()$ is a function that counts ones, $B_m$ is the set of different values for bits in mask $m$, and operator $\oplus*$ transforms mask to the schema. For $\alpha_{\{1,2\}}$, we have $M = \{1,2\}$, $m = [0\ 1\ 1]$ and $B_m = \{[0\ 1\ 1], [0\ 1\ 0], [0\ 0\ 1], [0\ 0\ 0]\}$.

### B. Walsh Coefficient metric for problem difficulty

A problem can be challenging for an EDA when there is a dependency between two variables but the estimated coefficient between them does not show an interaction. Another difficult situation occurs when coefficients between dependent variables and independent variables are similar. So a metric can be defined as

$$M_1 = \frac{|\alpha_D|-|\alpha_I|}{f_{max}}, \quad (20)$$

where $f_{max}$ is the maximum fitness value in the problem, $D$ is the set of two dependent variables and $I$ is the set of two independent variables. When M1 increases, the similarity between dependent and independent sets reduces, so the problem becomes easier for EDAs. To use this metric dependent and independent sets of variables in the problem should be given. When the fitness function itself is unavailable, this can not be achieved. Additionally, this metric cannot be used when the variables are all dependent on each other and there is no independent set, for example in one-max problem.

### C. Schema fitness estimation

Order two Walsh coefficients of the problem should be determined before using this metric. Walsh decomposition can be applied to simple problems with low dimensions. But for high dimension complex problems, the fitness values of schemas can be utilized to calculate the coefficient based on the formula. there are two methods to estimate $f_{a_i=x,a_j=y}$: confusion method and population-based method. In the confusion method, a random solution is produced; then the fitness values of 4 possible solutions by changing $i$th and $j$th bits are calculated, and using formula Walsh coefficient is estimated. in the population-based method, a random population of samples from the fitness function landscape is drawn, then the schema's fitness $f_{a_i=x,a_j=y}$ is estimated by their average fitness value. This estimation is more accurate with larger population size. The confusion method is faster with lower accuracy but the population technique finds a better estimation of the coefficient.

## V. EXPERIMENT

A problem is hard for an evolutionary algorithm if it needs a large number of fitness function calls. Thus, the proposed metric needs to be validated with regards to the fitness calls in order to determine whether it would be able to determine a problem's difficulty. For this purpose, Different EDAs are run on a couple of problems and the number of fitness function calls are recorded.

### A. Experiment conditions

We apply BOA and ECGA algorithms to different types of functions in varying dimensions. A question that arises here is how large the population of the algorithms should be. This question can be answered using the bisection algorithm. Evolutionary algorithms utilize the bisection method for finding the population size in order to solve problems.

---

1. Start with some small population N
2. Double N until algorithm convergence is sufficiently reliable
3. After algorithm convergence is reached, the minimum of adequate population is N/2 and maximum is N
4. Repeat until (max-min)/min is a percentage of N

---

Fig. 1. Bisection Algorithm

It is important to note that the population size discovered by this method is not so large that the problem would be easy for the algorithm or is not too small that the problem would be hard for the algorithm. So the population size would be optimum and as a result, the number of fitness calls is reasonable. The run configuration for BOA and ECGA is presented in table 2.

TABLE II. RUN CONFIGURATION BY BISECTION ALGORITHM

| | |
|---|---|
| **Initial Population Size** | 1000 |
| **Successful consecutive runs of bisection for convergence** | 50 |
| **Structure learning score** | BIC |
| **Maximum number of parents for each node** | 10 |

### B. Fitness Functions

In this section, A variety of fitness functions for running EDAs and evaluating the metric are introduced.

#### 1) K-Trap functions

This fitness function consists of a set of trap function[25] that is defined as

$$f_{\text{trap}(m,k)}(x) = \sum_{i=1}^{n/k} \text{trap}_k(u(x_{k(i-1)}, \ldots, x_{ki-1})) \quad (21)$$

$$\text{trap}_k(l) = \begin{cases} k & l = k \\ k - l - 1 & l < k \end{cases},$$

where $u()$ is the one max function that simply counts ones, $n$ is the length of the problem, $k$ is the length of each subfunction, and $m = n/k$ is the number of trap subfunction.

*2) Mixed Trap functions*

A mixed trap function switches between trap and inverse trap functions by a control bit and has two global optimums[26]. The inverse trap is defined as

$$f_{\overline{\text{trap}}(m,k)}(x) = \sum_{i=1}^{n/k} \overline{\text{trap}}_k(u(x_{k(i-1)}, \ldots, x_{ki-1})) \quad (22)$$

$$\overline{\text{trap}}_k(l) = \begin{cases} k & if\ l = 1 \\ l - 1 & otherwise \end{cases}.$$

The first and second mixed trap functions are

$$\text{trapI1}_k(x) = \begin{cases} f_{\overline{\text{trap}}(m,k)}(x_1, \ldots, x_{n-1}) & x_0 = 0 \\ f_{\text{trap}(m,k)}(x_1, \ldots, x_{n-1}) & x_0 = 1 \end{cases} \quad (23)$$

$$\text{trapI2}_k(x) = \begin{cases} H_0(x_1, \ldots, x_{n-1}) & x_0 = 0 \\ H_1(x_1, \ldots, x_{n-1}) & x_0 = 1 \end{cases} \quad (24)$$

$$H_0(x_1, \ldots, x_{n-1}) = f_{\overline{\text{trap}}(m,k)}(x_1, \ldots, x_{n-1})$$

$$H_1(x_1, \ldots, x_{n-1}) = \sum_{i=k/2}^{n/k-1} \text{trap}_k(u(x_{k(i-1)}, \ldots, x_{ki-1})) + \text{trap}_k(u(x_1, \ldots, x_{\frac{k}{2}}, x_{n-\frac{k}{2}}, \ldots, x_{n-1})).$$

The global optimums for both of these functions are full zero and full one bits.

*3) Multi-structure problems*

In multi-structure problems, the fitness of an input string is evaluated by two different structures[27]. in order to solve multi-structure problems, EDAs have to discover the interaction between variables as well as the interaction between variables. The first multi-structure problem is

$$MSP1_k(x) = \begin{cases} \alpha + f_{\text{trap}(m,k)}(x_1, \ldots, x_{n-1}) & x_0 = 1 \\ n - 1 - f_{onemax}(x_1, \ldots, x_{n-1}) & x_0 = 0 \end{cases}, \quad (25)$$

where $\alpha$ is the bias between two models. In this function, the global optimum is full one-bit string with a fitness value of $n + 1$ produced by the trap model and the local optimum is full zero-bit string with $n - 1$ fitness value by the one-max model. The optimum structure of this problem has interaction between the first bit and all other bits. The second multi-structure problem is

$$MSP2_k(x) = \max\{\alpha + f_{\text{trap}(m,k)}(x), n - 1 - f_{onemax}(x)\}. \quad (26)$$

In this function, the global optimum is full one-bit string produced by trap function with $n + 1$ fitness value and local optimum is full zero-bit string by one-max function with $n - 1$ fitness. With the removal of the control bit in this problem, EDA would have no idea about interactions and this would make the problem challenging. The third multi-structure problem is

$$MSP3_{k_1,k_2}(x) = \max\{\alpha + f_{\text{trap}(m_1,k_1)}(x), f_{\text{trap}(m_2,k_2)}(\bar{x})\}, \quad (27)$$

where $\bar{x} = 1 - x$ is the inverse of binary string $x$ and $k_1 \neq k_2$ and $m_1 k_1 = m_2 k_2$. In this function, there are interactions with different orders which can mislead linkage learning to discover the real structure of the problem.

We use these fitness functions to compare the ability of metrics to predict problem difficulty. Table 3 shows the configuration of different fitness functions for BOA and ECGA.

TABLE III. FITNESS FUNCTIONS CONFIGURATION FOR RUNS

| Problems | [Trap,TrapI1,TrapI2,MSP1,MSP2,MSP3] |
|---|---|
| n(BOA) | [30,45,60,75,90] |
| n(ECGA) | [12,15,21,27,30] |
| k | [3,5] |
| α | 1 |

*C. metrics*

The metric M1 is compared to mutual information and mutual entropy metrics proposed by Martins as

$$M_2 = I(X; Y) - I(X; Z) \quad (28)$$

$$M_3 = 1 - \frac{H(X, Y)}{H(X, Z)}, \quad (29)$$

and FDC as a traditional powerful metric is defined as

$$\text{FDC} = \frac{\frac{1}{n}\sum_{i=1}^{n}(f_i - \bar{f})(d_i - \bar{d})}{\sigma_F \sigma_D}. \quad (30)$$

For each function, dependent and independent variables are identified. Then a population of samples is drawn from the fitness function and one selection step is performed that keeps the best solutions and the metrics are calculated. These steps are repeated 50 times and the average value is considered for the metric.

*D. Experiment results*

Having run BOA and ECGA algorithms on different problems and calculated metrics, we can now discuss the metric's ability to predict. According to Table 3, there are 74 possible problems for BOA and 55 for ECGA based on configuration parameters. In fig. 2 and fig. 3 each sample represents a problem with one of the possible parameters; fitness calls count and the metric value are X and Y axis respectively. Since fitness calls count grows exponentially with problem difficulty, we use the logarithm of fitness calls as a representation of problem hardness. For some configuration parameters, BOA couldn't solve the problem and as a result, there is no fitness calls count available; so those problems are removed from the results which are MSP1 with $n = 75,90$ and MSP2 with $n = 45,60,75,90$.

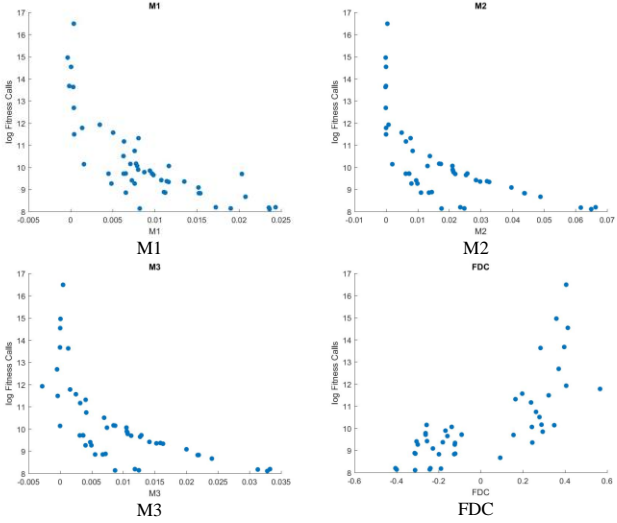

Fig. 2. BOA results for all metrics.

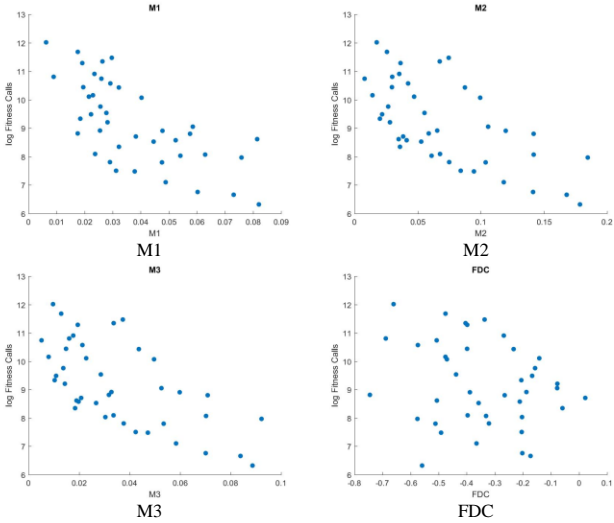

Fig. 3. ECGA results for all metrics.

To measure the association of each metric to problem difficulty, We use Pearson correlation coefficient, which shows the linear relation between metrics and fitness calls count. This coefficient ranges from -1 to 1. A value near 0 implies that there is no correlation, while a value near 1 or -1 shows correlation. Kendall's tau rank correlation is a nonparametric statistic used to measure the quality of ordinal relationship between two measured variables. Table 4 shows the Pearson and Kendall's tau coefficients between metrics and logarithm of fitness calls count.

TABLE IV. PEARSON AND KENDALL'S TAU COEFFICIENTS BETWEEN EACH METRIC AND LOGARITHM OF FITNESS CALLS

| Coefficient/Metric | M1 | M2 | M3 | FDC |
|---|---|---|---|---|
| Pearson(BOA) | -0.7320 | -0.6391 | -0.6321 | 0.7543 |
| Pearson(ECGA) | -0.7673 | -0.6961 | -0.6754 | 0.5732 |
| Kendall's tau(BOA) | -0.6959 | -0.6472 | -0.6481 | -0.2179 |
| Kendall's tau(ECGA) | -0.5455 | -0.5259 | -0.4785 | -0.1390 |

As shown in Table 4, absolute correlations for M1 are typically higher than other metrics, showing that the proposed metric has a better linear association to the problem hardness and can rank algorithms better than other metrics.

In fig 4. Kendall's value for total samples is shown as a red circle. If samples are grouped together by dimensions (which means all problems of each dimension is in one group) and the Kendall's value within each group is measured, a set of numbers is produced. The black line shows a 95 percent confidence interval of within group Kendall's values and the black circle is the mean.

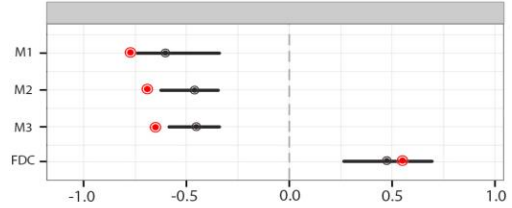

Fig. 4. Total Kendall's tau compared to within-group Kendall's tau.

Fig. 4 indicates that M1 has a high Kendall's value variance within groups which can be detrimental. Metrics M1, M2, and M3 all have lower mean within-group Kendalls meaning that they understand the problem gets harder as dimension increases, but they perform poorly in ranking functions within a dimension. Despite the lower total Kendall value, FDC has a better within-group mean and variance compared to M2 and M3 metrics.

Another way to compare predictions of these metrics is through linear regression. With this method, a linear model is constructed using M1, M2, and FDC metrics as input to calculate the fitness calls. Since M2 and M3 are both information-based metrics, to avoid shrinkage problem we removed M3 from this analysis.

$$y = \beta_0 + \sum_{i=1}^{n} \beta_i x_i \qquad (31)$$

In the learned model, the metrics' weights can represent their power to predict. The standardized regression coefficient of a variable, represents the expected change in output due to an increase in the variable with all other variables remaining unchanged. This means that the associated variable has a greater effect on the output when the coefficient increases. In Fig. 5 standardized M1 is higher than other coefficients in both algorithms which shows the regression model found M1 to be more correlated with fitness calls count.

TABLE V. COEFFICIENT OF DETERMINATION AND SIGNIFICANT IN THE LINEAR REGRESSION MODELS

|  | P | $R^2$ |
|---|---|---|
| BOA | <0.0005 | 0.613 |
| ECGA | <0.0005 | 0.641 |

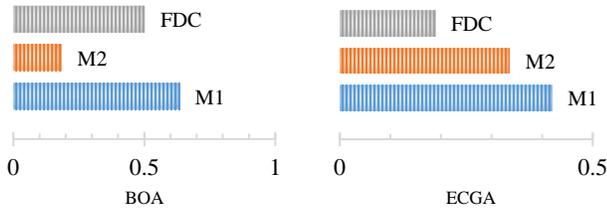

Fig. 5. Standardized Coefficients of metrics

Correlation coefficients, Kendall's coefficients, and linear regression models showed that M1 is better at predicting problem difficulty; Because after the selection step, they just use the entropy information of the remaining solutions and the fitness values information is lost. But M1 and FDC utilize this information. Furthermore, M1 and FDC do not require a selection step on drawn samples, but since M2 and M3 are information-based metrics and without a selection step samples have maximum entropy, they need this step.

We now can rank fitness functions by proposed metric. M1 is calculated for each function and the rank of functions is shown in the Fig. 6. According to M1, MSP2 is the most difficult problem and as mentioned before BOA couldn't solve it. Furthermore, The difficulty rises as k increases because interaction components get bigger and identifying the structure of the problem becomes harder.

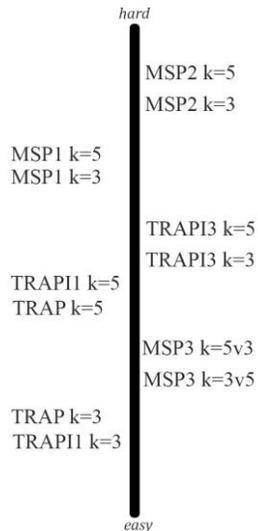

Fig. 6. Ranking of Fitness functions based on M1 metric

## VI. CONCLUSIONS AND FUTURE WORKS

Estimation of Distribution algorithms create a probabilistic graphical model and use drawn samples instead of crossover and mutation operators. The performance of these algorithms is highly dependent on the learned structure. They use pairwise statistics to model the relationship between variables in a problem. If these statistics do not show much difference between dependent and independent variables, the model building step of these algorithms can't find accurate models. In this study, we proposed a new metric based on Walsh coefficients of dependent and independent variables to measure problem difficulty in additively separable functions. We used this metric as an estimator of fitness functions call count. Results show that this metric can measure problem hardness for EDAs and predict their performance better than similar metrics. Walsh coefficients can be applied in future studies to measure the difficulty of problems with higher-order interactions between variables. Some problems involve multivariable interactions, but pairwise statistics fail to demonstrate them. In the future, it may be possible to use this metric and other metrics in models, such as neural networks or PGMs to estimate problem difficulty accurately.

REFERENCES

[1] G.Lu, J.Li and X.Yao, , Fitness Landscapes and Problem Difficulty in Evolutionary Algorithms: From Theory to Applications, Springer Berlin Heidelberg, 2014.

[2] J. He, T. Chen and X. Yao, "On the Easiest and hardest fitness functions," IEEE Transactions on Evolutionary Computation, vol. 19, no. 2, pp. 295-305, 2015.

[3] S. Baluja, "Population-Based Incremental Learning: A Method for Integrating Genetic Search Based Function Optimization and Competitive Learning," Technical Report, Pittsburgh, 1994

[4] G. R. Harik, F. G. Lobo and D. E. Goldberg, "The Compact Genetic Algorithm," IEEE TRANSACTIONS ON EVOLUTIONARY COMPUTATION, vol. 3, pp. 287-298, 1999.

[5] S. Yang, "Memory-enhanced univariate marginal distribution algorithms for dynamic optimization problems," in The 2005 IEEE Congress on Evolutionary Computation, 2005.

[6] S. Bonet, C. L. I. Jr and P. A. Viola, "MIMIC: Finding Optima by Estimating Probability Densities.," in Advances in neural information processing systems, Denver, 1996.

[7] M. Pelikan and H. Muehlenbein, "The Bivariate Marginal Distribution Algorithm," Advances in Soft Computing, pp. 521-535, 1991

[8] G. R. Harik, F. G. Lobo and D. E. Goldberg, "The Compact Genetic Algorithm," IEEE TRANSACTIONS ON EVOLUTIONARY COMPUTATION, vol. 3, pp. 287-298, 1999.

[9] M. Pelikan, D. E. Goldberg and E. Cantu-Paz, "BOA: The Bayesian Optimization Algorithm," in Genetic and Evolutionary Computation Conference (GECCO-99), 1999.

[10] J. H. Holland, Adaptation in Natural and Artificial Systems: An Introductory Analysis with Applications to Biology, Control and Artificial Intelligence, 1965

[11] D. E.Goldberg, Genetic Algorithms in Search, Optimization, and Machine Learning,, 1st, Ed., Boston: Addison-Wesley Longman, 1989.

[12] S. Writght, "The Roles of mutation, inbreeding, crossbreeding, and selection in evolution," in Proceedings of the 6th International Conference on Genetics, 1932

[13] E.-G. Talbi, Metaheuristics: From Design to Implementation, Chichester: Wiley, 2009.

[14] E. Weinberger, "Correlated and uncorrelated fitness landscapes and how to tell the difference," Biological Cybernetics, vol. 63, no. 5, pp. 325-336, 1990.

[15] P.Collard, S.Verel and M.Clergue, "Local search heuristics: Fitness Cloud versus Fitness Lancsape," ECAI, pp. 973-974, 2004.

[16] T. Jones and S. Forrest, "Fitness Distance Correlation as a Measure of Problem Difficulty for Genetic Algorithms," ICGA, pp. 184-192, 1995

[17] E. Weinberger, " Local properties of Kauffman's N-k model, a tuneably rugged energy," Phys Rev A., p. 6399–6413, 1991

[18] D. J. Coffin and R. E. Smith, "The limitations of distribution sampling for linkage learning," in Evolutionary Computation, 2007. CEC 2007. IEEE Congress on, 2007.

[19] C. Echegoyen, Q. Zhang, A. Mendiburu and R.Santana, "On the limits of effectiveness in estimation of distribution algorithms," in Proceedings of the Congress on Evolutionary Computation, New Orleans, 2011.

[20] J. P. Martins and A. C. Delbem, "Multimodality and the linkage-learning difficulty of additively separable functions," in Proceedings of the 2014

Annual Conference on Genetic and Evolutionary Computation, Vancouver, 2014.

[21] J. P. Martins and A. C. Delbem, "Pairwise independence and its impact on Estimation of Distribution Algorithms," Swarm and Evolutionary Computation, vol. 27, pp. 80-96, 2016

[22] D. E. Goldberg, "Genetic algorithms and Walsh functions: part I, a gentle introduction," Complex Systems, vol. 3, pp. 129-152, 1989.

[23] B. J. Fino and V. R. Algazi, "Unified matrix treatment of the fast WalshHadamard transform," IEEE Transactions on Computers, pp. 1142-1146, 1976.

[24] M. Munetomo and D. Goldberg, "A genetic algorithm using linkage identification by nonlinearity check," in *Systems, Man, and Cybernetics*, 1999.

[25] M. Pelikan and D. E. Goldberg, "Hierarchical problem solving and the Bayesian optimization algorithm," in GECCO'00 Proceedings of the 2nd Annual Conference on Genetic and Evolutionary Computation, San Francisco, 2000.

[26] C.-Y. Chuang and W.-L. Hsu, "Multivariate multi-model approach for globally multimodal problems," in *GECCO '10 Proceedings of the 12th annual conference on Genetic and evolutionary computation*, Portland, 2010.

[27] H. Sharifi, A. Nikanjam, H. Karshenas and N. Najimi, "Complexity of model learning in EDAs: multi-structure problems," in GECCO Comp '14 Proceedings of the Companion Publication of the 2014 Annual Conference on Genetic and Evolutionary Computation, Vancouver, BC, Canada, 2014.